\documentclass[a4paper,fleqn]{cas-dc}

\usepackage[authoryear]{natbib}
\usepackage[switch]{lineno}
\usepackage{enumitem}  
\usepackage{needspace}  
\usepackage[normalem]{ulem} 

\def\tsc#1{\csdef{#1}{\textsc{\lowercase{#1}}\xspace}}
\tsc{WGM}
\tsc{QE}
\tsc{EP}
\tsc{PMS}
\tsc{BEC}
\tsc{DE}

\newcommand{\revcolor}[1]{{\color{black} #1}}  
\usepackage{xcolor} 
\usepackage{hyperref} 
\usepackage{xurl}

\begin{document}
\let\WriteBookmarks\relax
\def\floatpagepagefraction{1}
\def\textpagefraction{.001}
\shortauthors{C-.T. Vu et~al.}

\title [mode = title]{
  Context-Aware Adaptive Pesticide Spraying for Agricultural Robots under Changing Weather and Terrain Using Vision–Language Models
}     


\author{{Cong-Thanh Vu}}
\ead{vuthanh.cdt@gmail.com}
\author{{Yen-Chen Liu}}\cormark[1]
\ead{yliu@mail.ncku.edu.tw}
\cortext[cor1]{Corresponding author}
\affiliation{organization={Department of Mechanical Engineering, National Cheng Kung University},
                city={Tainan},
                postcode={70101}, 
                country={Taiwan}}

\begin{abstract}
Precision pesticide spraying is essential for optimizing application efficiency and ensuring uniform chemical distribution. Spraying performance is influenced by multiple factors, including environmental conditions such as temperature and wind speed, pesticide type, and the robot’s capability to accurately perceive crops and target spray locations. Existing approaches predominantly emphasize crop detection and rely on predefined spraying parameters, whereas human operators dynamically adjust their spraying strategies by considering environmental conditions, region-specific crop characteristics, and the type of pesticide being applied. In this study, we propose a context-aware adaptive spraying framework based on Vision-Language Models (VLMs), which enables robots to leverage spatial reasoning and integrate information from multiple sources, including crop type, pesticide type, and weather data, to make adaptive and optimized spraying decisions. Subsequently, a trajectory-tracking controller based on Model Predictive Path Integral (MPPI) control is employed to ensure precise navigation and accurate spraying at crop locations. The comparative results demonstrate that the proposed method improves accuracy by at least 30\% in detecting crop rows. In addition, the experimental evaluations conducted in two environments further demonstrate the robot’s ability to flexibly adjust spraying volume and travel speed, while reducing pesticide drift.
\end{abstract}


\begin{keywords}
Precision spraying \sep Vision-language models \sep Agricultural automation
\end{keywords}

\maketitle

\section{Introduction}
Pesticide application has long been an essential component of agricultural production, helping protect crops from pests and diseases and reducing yield losses. However, current spraying practices, ranging from manual backpack spraying to autonomous systems, still present several limitations. During manual spraying, human operators can directly observe crop conditions and adjust spray volume, modify their movement direction based on wind or weather conditions, vary walking speed, or alter spraying patterns when encountering dense canopies, severely infected plants, or hard-to-reach areas. Although this flexibility is valuable, it depends heavily on operator experience, is physically demanding, and often results in inconsistent performance, which can negatively affect human health \citep{DASGUPTA2007121,SHEKHAR2024101840}. In contrast, mechanized and semi-automated sprayers lack such human-like adaptability; they typically apply pesticides continuously at fixed rates without accounting for spatial variability in canopy structure, leaf density, or disease severity \citep{GAO2025110960,FU2026111156,ZHANG2025110606,RAHIMIAZGHADI2025110365,7869323}. As a result, excessive pesticide deposition frequently occurs, increasing production costs, generating unnecessary chemical waste, and posing risks to both the environment and human health \citep{https://doi.org/10.1002/ps.5421}. These limitations underscore the need for more intelligent and adaptive spraying solutions that combine automated operation with decision-making capabilities resembling those of skilled human operators \citep{9122412,10459067}.

Over the past two decades, numerous studies have focused on developing automated systems to minimize pesticide waste and improve precision spraying. These systems aim to detect crops accurately and apply pesticides only where needed, rather than uniformly across the field. For example, \cite{NGUYEN2023108429} proposed a camera-based system that estimates tomato leaf density and selectively activates nozzles, overcoming the limitations of fixed spray trajectories, though the number of independently controllable spray positions remains limited. To improve coverage, \cite{fu2022improved} introduced adjustable nozzle heights for consistent deposition across varying plant structures. \cite{FU2026111156} further enhanced detection by compensating for camera tilt caused by uneven terrain. Despite these advances, vision-based systems struggle under low-light conditions; to address this, \cite{JIANG2025110359} integrated thermal imaging with LiDAR for robust crop detection in visually degraded environments.

Beyond crop detection for targeted spraying, weed detection and treatment have also attracted significant attention. \cite{10543078} developed a weeding platform that identifies weeds and generates spray plans to avoid damaging nearby crops. \cite{JIN2025110990} combined ultra-fine needle injectors with deep learning to treat individual weeds with high precision. Instead of detecting weeds directly, \cite{9830878} represented weed-infested areas as line segments and optimized nozzle allocation using binary linear programming to maximize coverage efficiency. Research on accurate target identification has also expanded to aerial platforms. \cite{luo2024extraction} used LiDAR point clouds to analyze canopy structures for drone spraying decisions, while \cite{WEI2024109425} employed YOLO-based fruit detection to enable tree-specific, highly localized pesticide applications.

Although crop detection and spraying decision-making systems have proven effective in reducing pesticide usage, several challenges remain. Variability in crop size and canopy structure makes maintaining a consistent spray rate difficult \citep{hussain2025enhancing, TASEER2024108841}. To address this, \cite{7869323} developed a device that adjusts flow rate and droplet size based on crop dimensions, but its reliance on color-based target detection and neglect of environmental factors limited its adaptability. Building on this, \cite{HU2025110379} proposed a multi-dimensional prescription map using 3D point clouds, incorporating leaf–wood separation via LSTM, nozzle position optimization, and spraying-unit division. RGB-D imagery has also been leveraged for real-time canopy segmentation and dynamic flow-rate adjustment \citep{GAO2025110960}, while \cite{ZHANG2025110606} fused camera and laser data to estimate canopy volumes for adjusting spray flow. Beyond flow control, regulating spray direction is crucial; \cite{shi2024development} introduced an active nozzle-orientation mechanism that aligns sprays with the canopy, minimizing off-target deposition and environmental impact.

Research on crop recognition and spray-rate adjustment has contributed significantly to reducing pesticide loss and mitigating environmental impact. However, these approaches still face many challenges, particularly in adapting spraying strategies flexibly to varying conditions, as a human operator would. In agricultural environments, farmers rely on contextual information, adjusting pesticide amounts based on variations in plant density, canopy structure, and disease status. They also consider weather conditions and the characteristics of each pesticide to determine appropriate movement directions and speeds, ensuring effective application according to temperature and wind. These factors, which are critical for spray coverage and droplet adhesion, have been largely overlooked in previous studies that primarily assume fixed movement trajectories. Therefore, developing an autonomous spraying system capable of making movement and spraying decisions in a human-like manner holds significant potential for advancing smart agriculture.

Recently, foundation models, such as Large Language Models (LLMs) and Vision-Language Models (VLMs), have achieved significant breakthroughs in human-like decision making and reasoning by leveraging large-scale Internet data. These models have been successfully applied in various robotics domains, including autonomous driving \citep{10531702}, natural-language-guided navigation using hand-drawn maps \citep{11027320}, and map-free outdoor navigation \citep{10960724}. They also show strong potential in agriculture, helping overcome the limitations of previous methods that were often tailored to specific problems and difficult to generalize \citep{11173627, KUSKA2024108924}. Several studies have explored these reasoning capabilities to enhance agricultural applications, such as multi-level tea bud detection with generative AI support during training, agriculture-specific model designs \citep{yang2025agrigpt}, and LLM-based analysis of data collected from agricultural environments \citep{11194249}.

In this study, we propose a novel approach for autonomous pesticide application using VLMs. The method enables the robot to execute \revcolor{a context-aware adaptive spraying strategy} based on weather conditions and the operational environment. The main contributions of this study are as follows:
\begin{itemize}[noitemsep]
\item We introduce a framework that integrates VLMs into autonomous pesticide spraying, enabling the robot to generate adaptive pesticide application strategies that take into account crop types, the operational environment, and external environmental factors such as wind direction and temperature.
\item We propose a crop-row detection method based on VLMs for global planning and compare it with other crop-row detection methods using unsupervised algorithms such as KMeans \citep{s23104808}, DBSCAN \citep{10773736}, or manual adjustments \citep{10604355}. The proposed approach significantly improves crop-row detection accuracy by at least 30\%.
\item We develop a path-tracking controller based on the Model Predictive Path Integral (MPPI) framework for pesticide spraying tasks, enabling the robot to follow the spraying trajectory while simultaneously avoiding obstacles during navigation.
\item The proposed method is evaluated in two different field environments. The results show that our approach generates \revcolor{context-aware adaptive spraying strategies} under varying environmental conditions and also helps reduce pesticide loss caused by environmental factors.
\end{itemize}
The remainder of this paper is organized as follows. Section \ref{sec:system} describes the system design, including the hardware configuration and component connectivity. Section \ref{sec:method} presents the proposed methodology. The experimental setup, results, and discussions are provided in Section \ref{sec:results}. Finally, Section \ref{sec:conclusion} summarizes the key findings and future work.
\section{System Design}\label{sec:system}
\begin{figure*}[t]
\centering
\includegraphics[scale=0.83,page=1]{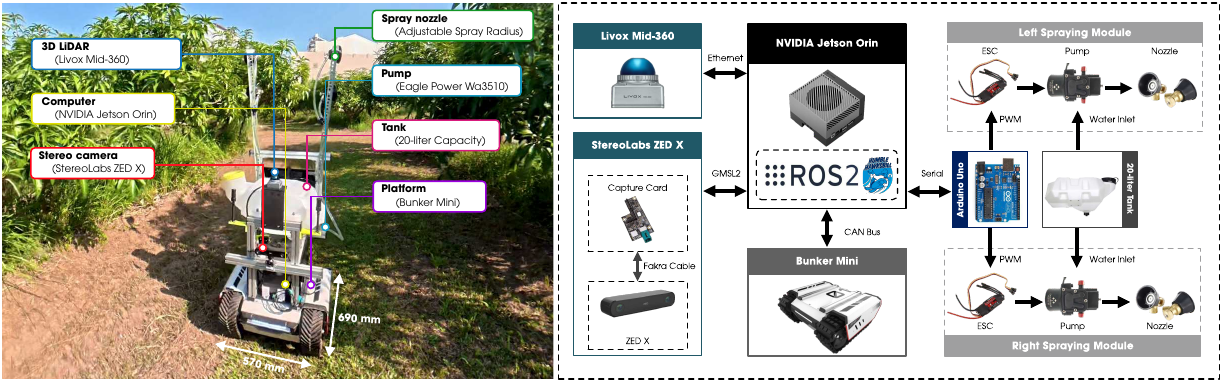}
\caption{Hardware configuration of the pesticide-spraying robot. The LiDAR sensor supports localization and mapping, while the camera enables crop recognition. The NVIDIA Jetson Orin–based central processor handles perception, generates movement commands for the mobile base, and controls the two spraying modules on either side of the robot.}
\label{fig:hardware}
\end{figure*}
This section presents the autonomous pesticide-spraying robot used in the experiments. The robot is developed based on the Bunker Mini platform manufactured by Agilex. This compact chassis employs a tracked mobility system, enabling stable operation on uneven and complex terrains commonly found in agricultural fields, as illustrated in Fig.~\ref{fig:hardware}.

At the front of the robot, we integrate a ZED X camera connected to the Jetson Orin onboard computer via a GMSL2 interface, along with a Livox Mid-360 3D LiDAR connected through Ethernet to support crop detection and spatial mapping. All sensor data processing and execution of the proposed algorithms are performed on the NVIDIA Jetson Orin edge computing device, which provides real-time performance through CUDA-accelerated parallel computation. The control framework is developed based on ROS2 Humble. The robot communicates with the Jetson Orin via a CAN bus to receive velocity commands and transmit status information.

For the spraying system, we design two independent modules mounted on the left and right sides of the robot, enabling side-specific spraying decisions instead of relying on a single central pump. This design helps optimize chemical usage and reduce waste. To control the two pump modules, an Arduino Uno is used as an intermediary controller, communicating with the Jetson Orin through Serial to receive pump commands. The Arduino then converts these commands into PWM signals to drive the ESCs and regulate the pump speed. The selected pump is the Eagle Power Wa3510, a compact BLDC motor capable of generating high pressure. Each side is equipped with two adjustable nozzles, allowing the spray diameter to be tuned for optimal chemical usage and coverage under varying field conditions. These pumps are connected to a 20-liter pesticide tank mounted on the robot chassis.
\section{Methods}\label{sec:method}
In this section, we present the proposed methodology, including the system architecture, the VLM-enabled path-planning module for autonomous crop row detection, the VLM-based spraying strategy, and finally, the path-tracking controller
\subsection{Architecture Overview}
The first step in our approach is to construct a global map of the field environment from point cloud data captured by a 3D LiDAR with a built-in IMU using the Fast-LIO2 algorithm~\citep{9697912}. During the mapping process, the robot pose is continuously estimated and represented in the global frame $\mathcal{G}$ as $\mathbf{q}^r = [x^r, y^r, \theta^r]^{\mathrm{T}} \in \mathbb{R}^3$. In parallel, the system employs a crop-detection module based on YOLOv8 \citep{Jocher_Ultralytics_YOLO_2023} with a ZED X camera, leveraging the computational acceleration provided by Tensor Cores on the Jetson Orin to identify trees and extract their positions in the robot’s local coordinate frame $\mathcal{R}$, with each tree position represented as $\mathbf{q}^r_{t,i} = [x^r_{t,i}, y^r_{t,i}, \theta^r_{t,i}]^{\mathrm{T}} \in \mathbb{R}^3$, for $i = 1,\ldots,n$, where $n$ is the total number of detected trees and $i$ denotes the unique identifier (ID) assigned to each tree. The transformation of coordinates between the global and local frames for tree positions, which is stored in the working map, can be expressed as:
\begin{align}
\mathcal{W} = \left\{ \mathbf{q}^t_{i} \mid i = 1, \ldots, n \right\}, \quad
\mathbf{q}^t_i = {^{\mathrm{G}}_{\mathrm{R}}{\mathbf{T}}}\mathbf{q}^r_{t,i}
\end{align}
where
\begin{align*}
   {^{\mathrm{G}}_{\mathrm{R}}{\mathbf{T}}} = \begin{bmatrix}
        \mathbf{J}(\theta^r) & \mathbf{q}^r \\
        0 & 1
    \end{bmatrix}, \quad
    \mathbf{J}(\theta^r) = \begin{bmatrix}
        \cos\theta^r & -\sin\theta^r & 0 \\
        \sin\theta^r & \cos\theta^r & 0 \\
        0 & 0 & 1
    \end{bmatrix},
\end{align*}
$^{\mathrm{G}}_{\mathrm{R}} \mathbf{T}$ is the transformation matrix from $\mathcal{R}$ to $\mathcal{G}$, and $\mathbf{J}(\theta_r)$ is the rotation transformation matrix.

Following the construction of a working map integrated with the global positions of individual trees, we propose a VLM-based tree row recognition method. This method uses a 2D representation of tree locations as input. By leveraging a prompt-based technique, the VLM accurately identifies the tree rows within the environment. Once the tree rows are recognized, a global path is generated along the identified rows. Subsequently, this path, together with the row-classified tree positions represented in 2D space and the real-world environmental images captured by the camera, is fed back into the VLM as input. In addition to the raw environmental data, supplementary information such as weather conditions and pesticide type is incorporated to provide richer contextual cues for the VLM. Based on the predicted route, weather conditions, pesticide type, and tree-related information, the VLM determines the appropriate travel velocity and spraying flow rate. This approach enables the spraying robot to operate flexibly in a human-like manner, simultaneously considering factors such as weather, chemical type, and terrain characteristics to derive a rational spraying strategy.

\begin{figure*}[t]
\centering
\includegraphics[scale=0.74,page=1]{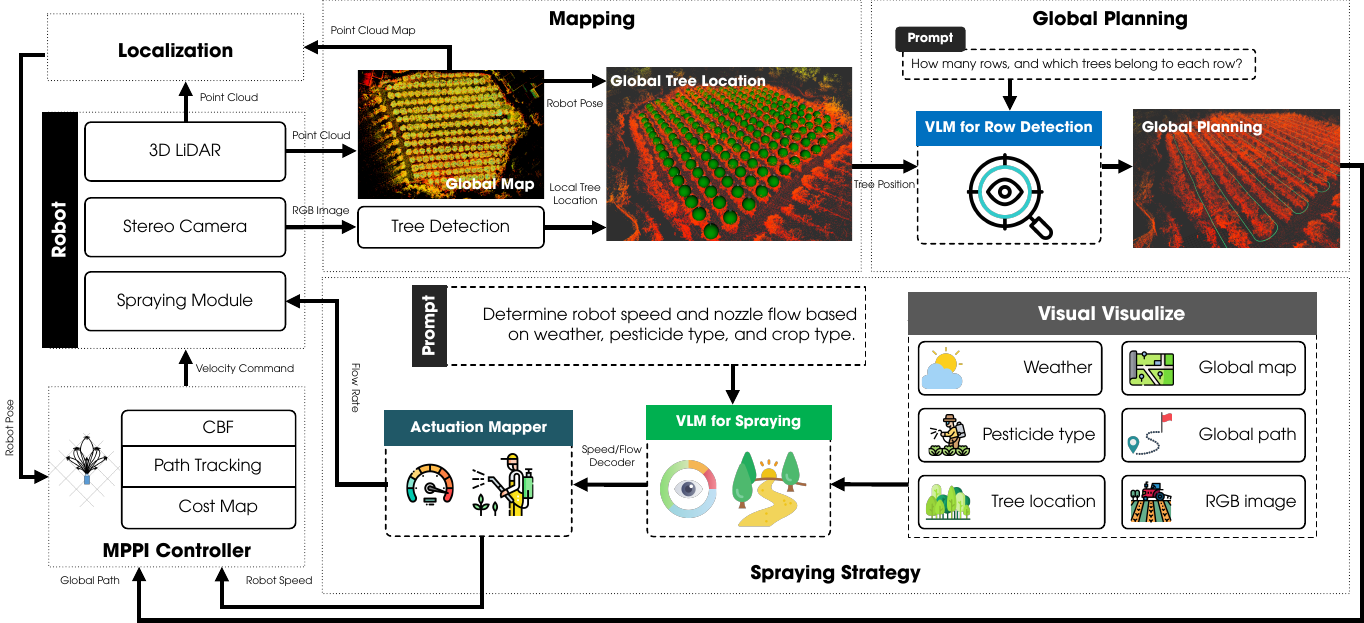}
\caption{Architecture overview of the proposed method.}
\label{fig:architecture}
\end{figure*}

The next stage involves an MPPI-based path-following controller, which is employed to track the global path. The robot’s state for the controller is determined using the constructed working map, the tree map, and local LiDAR data. This controller is designed to ensure that the robot adheres closely to the planned trajectory while actively avoiding obstacles and preventing collisions with crops. The overall architecture of the proposed methodology is illustrated in Fig.~\ref{fig:architecture}.
\subsection{VLM-Based Global Path Planning}\label{sec:row_detection}
\begin{figure}
\centering
\includegraphics[scale=0.193,page=1]{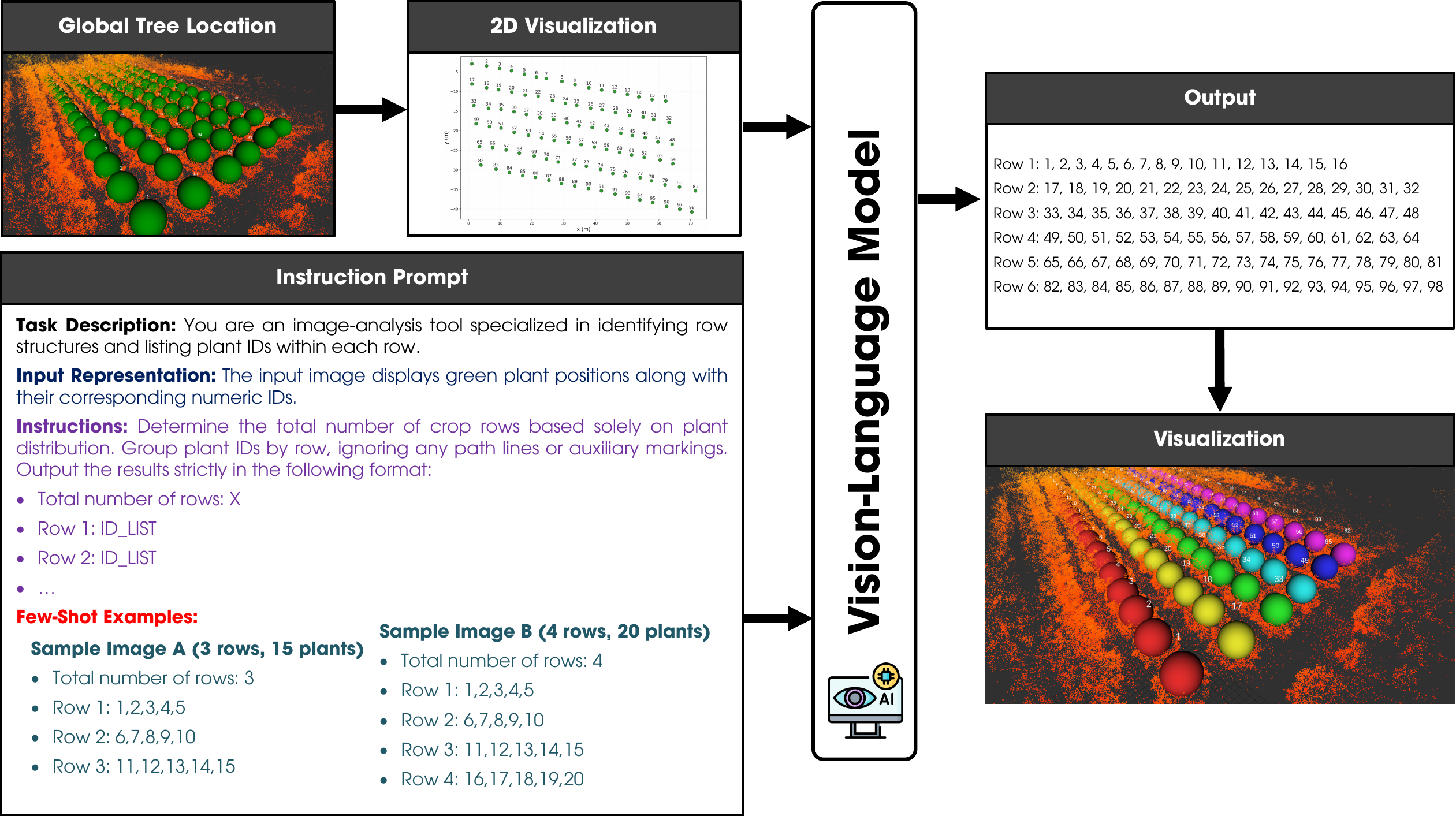}
\caption{Row classification using a VLM.}
\label{fig:row_detect}
\end{figure}
In the task of pesticide spraying, the robot needs to recognize crop rows during navigation to determine the appropriate path along each row. However, previous studies have often been limited in row identification, relying solely on the distances between plants along rows and columns through clustering algorithms using unsupervised learning \citep{10773736, s23104808} or manually defined parameters to reduce misclassification \citep{10604355}. In practice, especially in small-scale or unstructured agricultural environments, crops are not always evenly spaced, making distance-based grouping challenging and prone to errors in row classification. In contrast, humans can accurately identify and classify crop rows based on visual observation alone, without knowing the precise positions of individual plants. To address this limitation, we propose a method based on a VLM for crop-row classification, leveraging its ability to process diverse data and perform high-level semantic reasoning, similar to human cognition, to achieve more accurate row identification.

Therefore, the trees are represented in a 2D space and include both the coordinates and the IDs of individual trees. Instead of relying on the full 3D point cloud, which contains substantial redundant information and noise, we preserve only the essential spatial information of the trees. This compact representation enables the VLM to focus on the orchard row structure and the key spatial relationships, thereby enhancing its reasoning capability without being influenced by irrelevant components. The 2D image generated after preprocessing is used as the input to the VLM, where all tree coordinates are maintained to help the model accurately interpret the structure and relative arrangement of the trees. In addition, the image includes both the tree positions and their IDs, enabling the model to identify tree rows based on known IDs rather than predicting coordinates directly, which is a difficult and error-prone task for VLMs. Formally, the tree rows inferred by the VLM can be expressed as a set of row clusters $G$:
\begin{align}
G = \text{VLM}_{\text{row}}(\mathcal{I}_{\text{row}}, \mathcal{P}_\text{row}),
\end{align}
where $\mathcal{I}_{\text{row}}$ is the reconstructed visual representation of the tree positions annotated with their IDs, and $\mathcal{P}_\text{row}$ is a task-specific prompt that provides contextual information and guides the model to produce the desired output, as illustrated in Fig.~\ref{fig:row_detect}.
\begin{figure}
\centering
\includegraphics[scale=0.176,page=1]{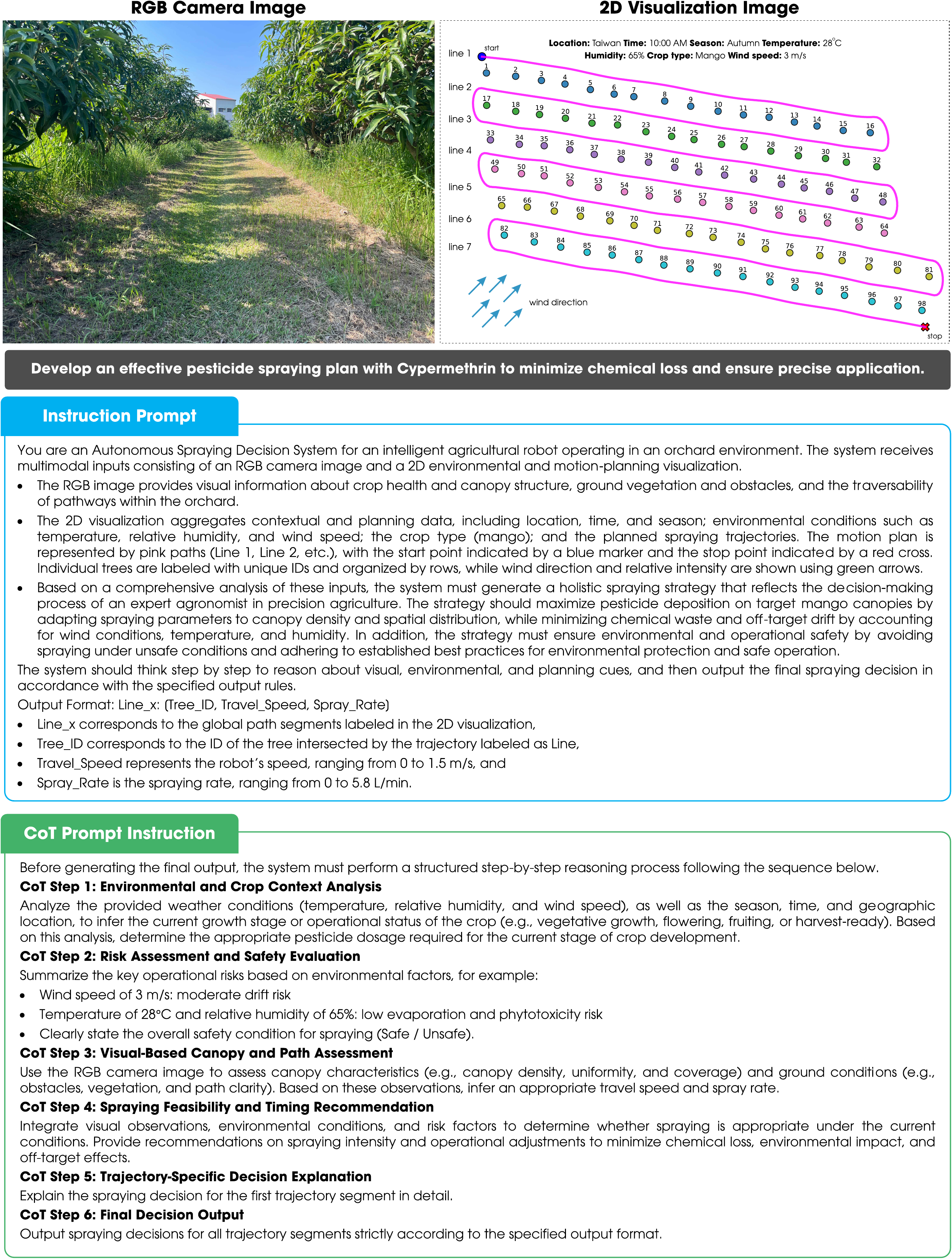}
\caption{Pesticide spraying strategy using VLM for autonomous agricultural robots.}
\label{fig:spraying_vlm}
\end{figure}
Rather than relying solely on images and conventional zero-shot prompts, we adopt prompt engineering in combination with few-shot prompting to guide the VLM toward consistent and accurate detection of crop rows. Specifically, we first explicitly define the task, ensuring that the VLM recognizes it must identify tree rows in the image based on the provided IDs, and we describe the information encoded within the image to guarantee correct interpretation of the input. Subsequently, we establish a set of rules that the VLM must adhere to, including requirements that all IDs in the output correspond to the provided dataset and that the output strictly conforms to the prescribed format. Finally, illustrative examples and output-format specifications are incorporated into the prompt to ensure that the model generates results that are rigorously constrained by the input data and fully compliant with the desired output structure.

After the corresponding crop rows are identified, centerlines are generated between adjacent rows, positioned equidistantly from the trees on either side to allow the robot to navigate along the middle. These centerlines are then sequentially connected to form a continuous path, enabling the robot to traverse the orchard in a systematic zig-zag pattern, efficiently covering the entire area.
\subsection{VLM-Based Spraying Strategy}
Previous studies have primarily relied on crop location information and pre-defined working maps to design rigid control strategies for pesticide spraying. Such approaches lack consideration of critical contextual factors, including crop growth stages, weather conditions, and other environmental variables that human operators typically evaluate during spraying operations. To address these limitations, the proposed method adopts a VLM-based framework that leverages human-like contextual reasoning and integrates visual perception with prompt engineering, as illustrated in Fig.~\ref{fig:spraying_vlm}.

Specifically, the VLM takes two types of visual inputs: RGB images captured by the onboard camera and a constructed 2D visual representation that provides enriched contextual information. The RGB images enable the VLM to infer environmental characteristics such as uneven terrain, which may affect robot locomotion, induce instability, and alter the spraying angle. Based on this perception, the VLM can make informed decisions on adjusting the robot’s traveling speed to maintain stability and spraying accuracy. Moreover, RGB images allow the model to perceive real-field conditions, including canopy development and surrounding vegetation growth, thereby facilitating context-aware spraying decisions.

In addition to RGB images, a 2D visual representation is introduced to further enrich the VLM input. This representation encodes crop locations using distinct colors for each classified crop row and overlays a global path indicating the planned robot trajectory for spraying. Notably, wind direction information obtained from the Windy API is also embedded in the 2D visualization, illustrating how wind affects spray dispersion as the robot moves along the global path. This enables the VLM to adaptively adjust traveling speed and spraying volume for individual crop rows in a manner analogous to human operators.

Furthermore, the VLM is provided with auxiliary semantic information, including crop type, geographical cultivation location, current season and time, as well as the type of pesticide being applied. In particular, by knowing the pesticide type, the VLM can determine the appropriate spraying strategy based on current weather conditions, RGB images from the real-world environment, and suitable spraying locations, such as whether to focus more on the canopy or the trunk. These inputs allow the model to assess potential spraying risks and adjust spraying volume according to crop-specific characteristics, as the same crop species may exhibit different cultivation properties across geographical regions. Additional environmental factors, such as humidity, temperature, and current wind speed, are also incorporated, enabling the VLM to evaluate whether prevailing conditions are suitable for pesticide spraying and to modify the spraying strategy accordingly.

To effectively leverage this diverse set of multimodal information, we employ a chain-of-thought (CoT) prompt design in place of the few-shot prompting strategy used in the crop row detection task. The CoT prompting mechanism enables the VLM to perform step-by-step reasoning in a human-like manner before producing the final decision. The pesticide spraying strategy based on the VLM can be expressed as:
\begin{align}
(speed, volume) = \text{VLM}_{\text{spray}}(\mathcal{I}_{\text{spray}}, \mathcal{P}_{\text{spray}}),
\end{align}
where $(speed, volume)$ correspond to the robot traveling speed and the spray nozzle flow rate, respectively. Here, $\mathcal{I}_{\text{spray}}$ denotes the input visual information, and $\mathcal{P}_{\text{spray}}$ represents the guiding prompt.

To ensure that the outputs of the VLM are compatible with the agricultural robot, we define a set of output rules that the VLM must follow in order to prevent the generation of inappropriate or unsafe commands. The VLM outputs are constrained to follow the spatial structure of the global path, such that each crop row and each plant along the path is assigned a unique identifier. As the robot traverses the global path, each identified plant is controlled with a corresponding traveling speed and spraying volume determined by the VLM.

The spraying decision sent to the spraying module is determined based on the distance between the robot and each plant. Specifically, when the Euclidean distance between the robot and a plant falls within a predefined threshold and the plant lies on the appropriate spraying side of the robot, the system triggers the spraying action using the volume level provided by the VLM. Meanwhile, the movement speed command is forwarded to the path following controller, as described in Section~\ref{controller}.
\subsection{Path Following Controller}\label{controller}
In challenging agricultural environments, which are highly nonlinear and involve significant dynamics due to uneven terrain and disturbances affecting the robot's tracking performance, we employ the MPPI controller \citep{8558663} to track the global path. MPPI handles such disturbances through its stochastic nonlinear MPC framework, making it well-suited for highly dynamic environments such as agriculture. Moreover, we leverage the parallel computing capabilities of the CUDA provided by the Jetson Orin to accelerate the computations required by MPPI. The controller aims to determine the optimal control inputs that minimize a designed cost function over a prediction horizon of $K$ discrete time steps, starting from the current step $k$.

The robot platform considered in this study is non-holonomic. Therefore, its discrete-time state for MPPI is defined as:
\begin{align}
    \mathbf{q}_{k+1}^r = \mathbf{q}_{k}^r + \mathbf{J}(\theta^r_k)\, \mathbf{v}^r_k \, \Delta t, \hspace{0.5em}
    \mathbf{J}(\theta^r_k) = 
    \begin{bmatrix}
    \cos\theta^r_k & 0 \\
    \sin\theta^r_k & 0 \\
    0              & 1
    \end{bmatrix},
\end{align}
where $\mathbf{q}^r_k$ denotes the robot state at time step $k$, and $\mathbf{v}^r_k = [v^{r}_k,  \omega^r_k]^{\mathrm{T}}$ is the control input applied at time step $k$, for $k = 0, \dots, K-1$, with sampling period $\Delta t$. Here, $v^{r}_k$ represents the robot's linear velocity, while $\omega^r_k$ denotes its angular velocity in the frame $\mathcal{R}$. The matrix $\mathbf{J}(\theta^r_k)$ is the rotation matrix that maps the control inputs to the state update for the non-holonomic platform.

The cost function, designed to optimize the global path tracking task, is defined as:
\begin{align}
C(\mathbf{q}^r_k) = C_{\text{tracking}}(\mathbf{q}^r_k) + C_{\text{obs}}(\mathbf{q}^r_k) + C_{\text{vel}}(\mathbf{q}^r_k),
\end{align}
where $C_{\text{tracking}}(\mathbf{q}^r_k)$ promotes the robot's adherence to the global path, $C_{\text{obs}}(\mathbf{q}^r_k)$ facilitates obstacle avoidance, and $C_{\text{vel}}(\mathbf{q}^r_k)$ allows the robot to move at the speed provided by the VLM. The tracking cost can be expressed as:
\begin{align}
C_{\text{tracking}}(\mathbf{q}^r_k) &=  w_p \left\|\mathbf{q}^r_k - \mathbf{q}^l_k\right\| + w_n \left\|\mathbf{q}^r_k - \mathbf{q}^n_k \right\|,
\end{align}
where $\mathbf{q}^l_k$ is the lookahead point on the global path at a distance $l$ ahead of the robot, allowing the robot to progress forward along the path, while $\mathbf{q}^n_k$ is the nearest point on the global path to the robot's current position, used to maintain lateral alignment with the path. Here, $w_p$ and $w_n$ are the corresponding weights.

The obstacle avoidance cost allows the robot to navigate without collisions based on observations from the environment using a 3D LiDAR. To enhance computational efficiency, we utilize the spatio-temporal voxel layer costmap \citep{stvl} instead of directly evaluating each point from the raw point cloud data, where $l(\mathbf{q})$ denotes the function used to determine the cost at a given point $\mathbf{q}$. The obstacle avoidance cost is defined as:
\begin{align}
C_{\text{obs}}(\mathbf{q}^r_k) = w_{o} \begin{cases}
c_{\text{collision}}, & \text{if } l(\mathbf{q}^{r}_{k}) \ge \tau, \\
0, & \text{otherwise.}
\end{cases}
\end{align}
Here, the weighting factor $w_{o}$ corresponds to obstacle avoidance, and $\tau$ is a threshold used to detect potential collisions based on the local costmap.


The movement speed cost, $C_{\text{vel}}$, which allows the robot to track the speed provided by the VLM is defined as:
\begin{align}
C_{\text{vel}}(\mathbf{q}^r_k) = w_v \, \| v_k^r - v_{\text{vlm}} \|,
\end{align}
where $v_{\text{vlm}}$ is the desired speed from the VLM, and $w_v$ is the corresponding weight.

\revcolor{The MPPI controller is implemented using the Numba library and leverages CUDA parallel computation, operating at a frequency of 100 Hz. The parameters for this module are set as follows: $K = 3.0$, $\Delta t = 0.1$, $\tau = 50$, $w_p = 20.0$, $w_n = 5.0$, $w_o = 1000.0$, $w_c = 1000.0$, $w_v = 20.0$.} 
\section{Experiments Results and Discussion}\label{sec:results}
This section presents the experimental evaluation of the proposed method, including the assessment of crop detection for global path planning, the evaluation of the VLM-based spraying strategy in two real-world environments, and the performance of the controller. Finally, the results are discussed to highlight key insights and potential areas for improvement.
\subsection{Crop Row Detection Evaluation}
To evaluate the capability of tree-row detection for global path planning in robotic systems, we collected crop distribution data from 20 different environments, in which tree rows are arranged according to configurations commonly observed in real-world farms. Based on this dataset, the proposed tree-row detection method is compared with related studies in \cite{10773736, s23104808}, which employ unsupervised clustering algorithms such as K-means and DBSCAN for tree-row detection. In addition, we evaluate the performance of six VLMs, including Gemini 2.0 Flash Lite, Gemini 2.5 Flash Lite, Gemini 2.5 Flash, GPT-4.1 Nano, GPT-4.1 Mini, and GPT-4.1, on the tree-row detection task. Each experimental scenario is repeated 10 times for each method to ensure objectivity and reliability of the results.

\begin{figure}
\centering
\includegraphics[scale=0.3,page=1]{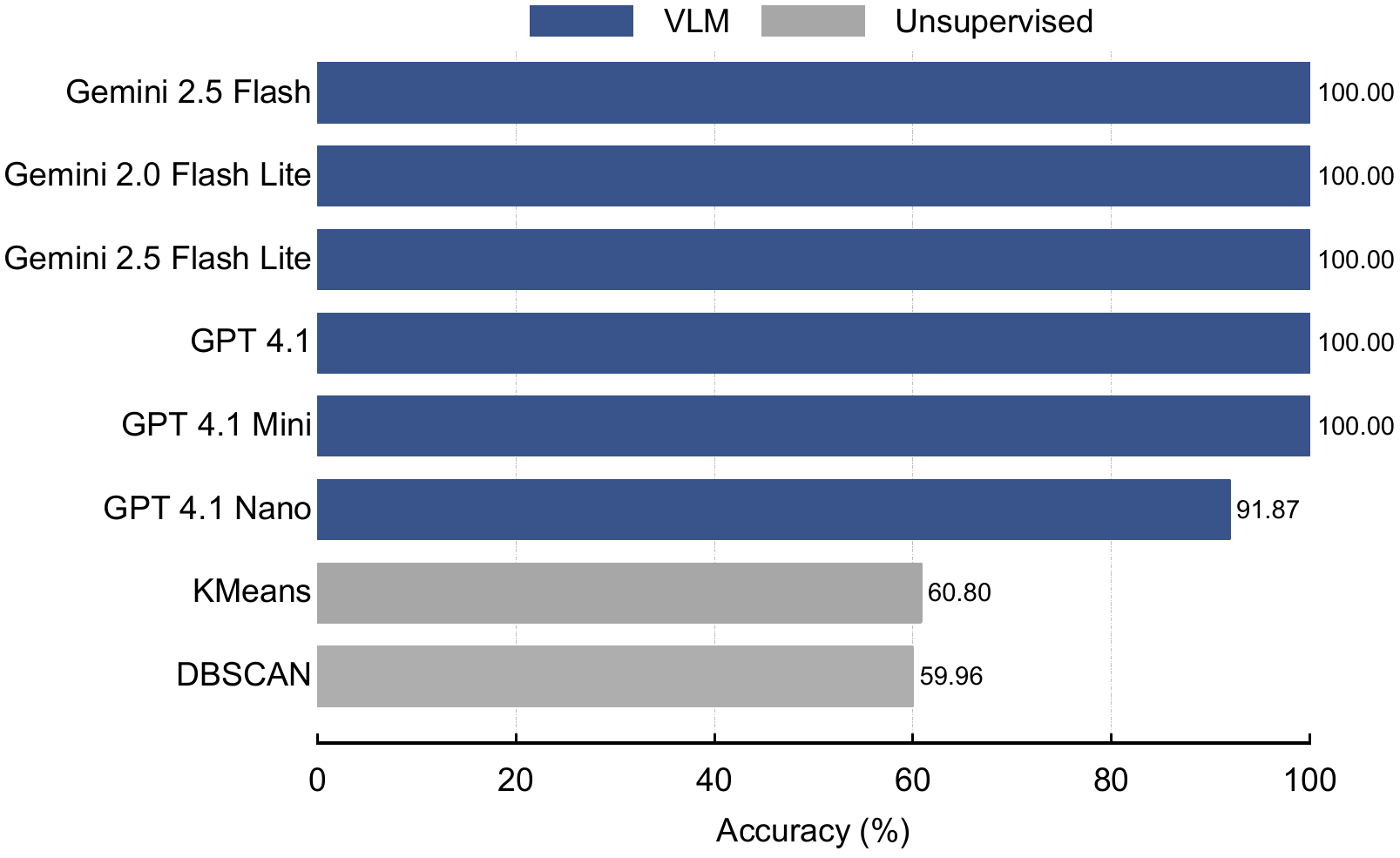}
\caption{Comparison results of tree-row detection.}
\label{fig:row_detection_results}
\end{figure}

The accuracy is defined as the number of correctly identified trees belonging to tree rows. The comparative performance of the VLM-based approaches versus traditional unsupervised learning methods is presented in Fig.~\ref{fig:row_detection_results}. The experimental results show that the VLMs demonstrate clear superiority in accuracy over traditional clustering-based methods. Specifically, Gemini 2.5 Flash, Gemini 2.0 Flash Lite, Gemini 2.5 Flash Lite, GPT-4.1, and GPT-4.1 Mini all achieve high accuracy. VLMs are capable of recognizing row structures and understanding spatial context, even in their lightweight variants (Lite/Mini). In contrast, GPT-4.1 Nano, the smallest model in the GPT family, exhibits a reduced accuracy of $91.87\%$. Although lower than that of the larger models, this performance still significantly surpasses that of traditional methods, highlighting the potential of compact VLMs for reliable tree-row detection.

It is also notable that clustering algorithms such as K-means and DBSCAN, used for crop-row recognition, yield substantially lower accuracies of $60.80\%$ and $59.96\%$, respectively. This performance gap (approximately 30–40\%) suggests that traditional methods are more sensitive to noise and complex row structures, whereas VLMs exhibit superior generalization from visual data. In addition, clustering-based approaches require manual tuning of parameters, such as the number of rows and inter-row distances, which limits their flexibility in diverse real-world environments. By contrast, the proposed method achieves high accuracy and enables automatic recognition of crop rows in agricultural environments, eliminating the need for manual adjustment as presented by \cite{10604355}.
\begin{figure}
\centering
\includegraphics[scale=0.29,page=1]{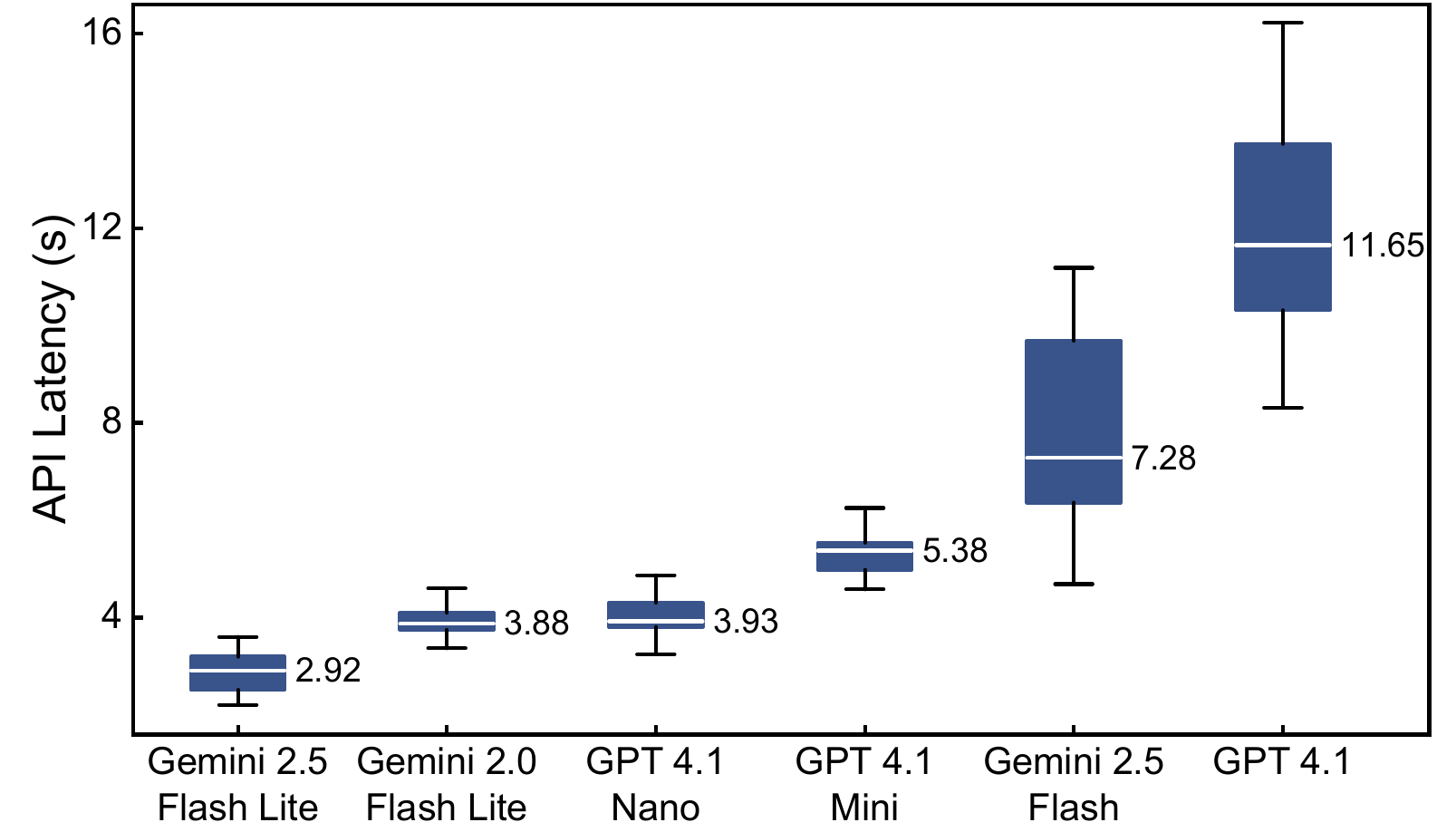}
\caption{API response latency of the compared VLMs.}
\label{fig:api_time}
\end{figure}

In addition to accuracy, we evaluate the average response latency of the models within the proposed tree-row detection framework, as shown in Fig. \ref{fig:api_time}. The results indicate that the Lite and Nano variants offer clear advantages in response time. In particular, Gemini 2.5 Flash Lite achieves the lowest average latency ($2.92$ s), followed by Gemini 2.0 Flash Lite ($3.88$ s) and GPT-4.1 Nano ($3.93$ s). GPT-4.1 Mini provides a reasonable trade-off between accuracy and computational efficiency, with an average latency of $5.38$ s, while the more computationally intensive models, Gemini 2.5 Flash and GPT-4.1, exhibit significantly higher latencies ($7.28$ s and $11.65$ s, respectively), with GPT-4.1 also showing greater variability in response time.

\revcolor{In the proposed system, tree-row detection is performed only during the global planning stage, where high-frequency execution is not required. Therefore, VLMs such as GPT and Gemini can still provide accurate recognition despite their relatively slower response times. In this study, Gemini 2.5 Flash was selected as the most suitable model because it offers high recognition accuracy while maintaining efficient inference performance.}
\subsection{Spraying Strategy Evaluation}\label{sec:spraying_exp}
{\subsubsection{Experimental Setup}
To evaluate the effectiveness of the proposed method under real-world conditions, two distinct field environments, shown in Fig.~\ref{fig:env_exp}, were selected and tested using the robotic platform described in Section~\ref{sec:system}. The first environment is a woodland area characterized by uneven tree density. To emulate orchard-like agricultural conditions, a section of this woodland was selected for its row-like vegetation structure, covering a total surveyed area of approximately 8,000 m$^2$.

The second environment is a mango farm covering approximately 5,000 m$^2$. This environment features clearly defined tree rows but is significantly affected by uneven terrain; the rows vary in length and alignment, posing substantial challenges for both perception and navigation.

\revcolor{During the field experiments, wind data acquisition was conducted using a hybrid approach to overcome the individual limitations of the measurement tools. Specifically, while the Windy API provides reliable macro-level wind direction, it lacks the localized precision required for accurate micro-climate wind speed estimation near the ground. Conversely, the onboard anemometer precisely measures real-time wind speed but lacks directional sensing capabilities. By combining these two sources, the macro-level wind direction from the Windy API was used to inform the overall spraying strategy, while the localized wind speed, which directly affects spray drift and flow rate adjustments, was continuously monitored by the anemometer. The sensor was mounted on the robot at a height of 1.67 meters, aligned with the elevation of the spray nozzles to accurately capture the aerodynamic conditions directly interacting with the sprayed droplets.}

Three different time points were selected in each environment to evaluate the adaptability of the proposed method to changing conditions, such as variations in wind direction and temperature. It should be noted that the robot’s traveling speed and spraying flow rate were automatically determined based on the proposed method and were not manually set by the user. \revcolor{The VLM model used in this work is Gemini 2.5 Flash.}
\subsubsection{Experimental Results}
\begin{figure*}
\centering
\includegraphics[scale=0.71,page=1]{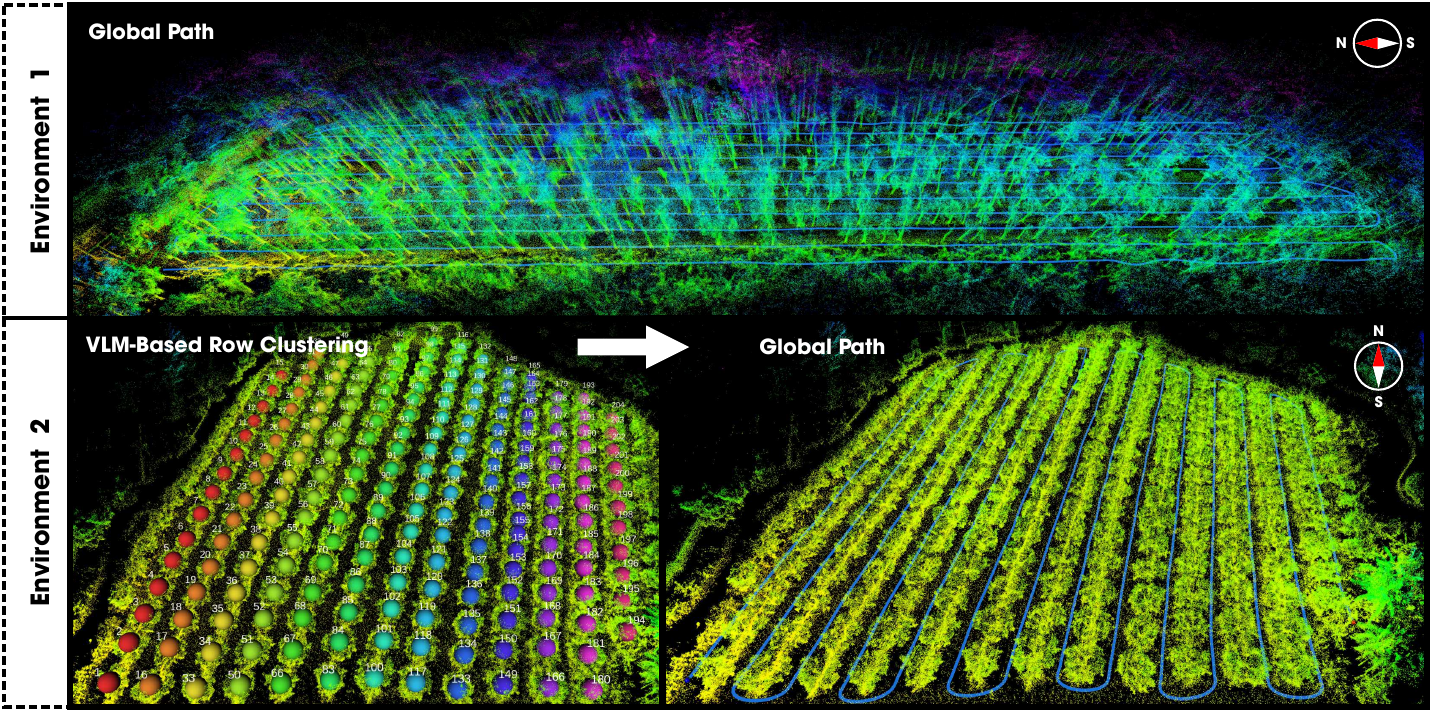}
\caption{Global maps and global paths of the two environments. The blue lines represent the global paths, while the labeled spheres indicate the detected tree positions. The colors of the spheres denote row membership as determined by the VLM.}
\label{fig:global_path}
\end{figure*}
\begin{figure}
\centering
\includegraphics[scale=0.325,page=1]{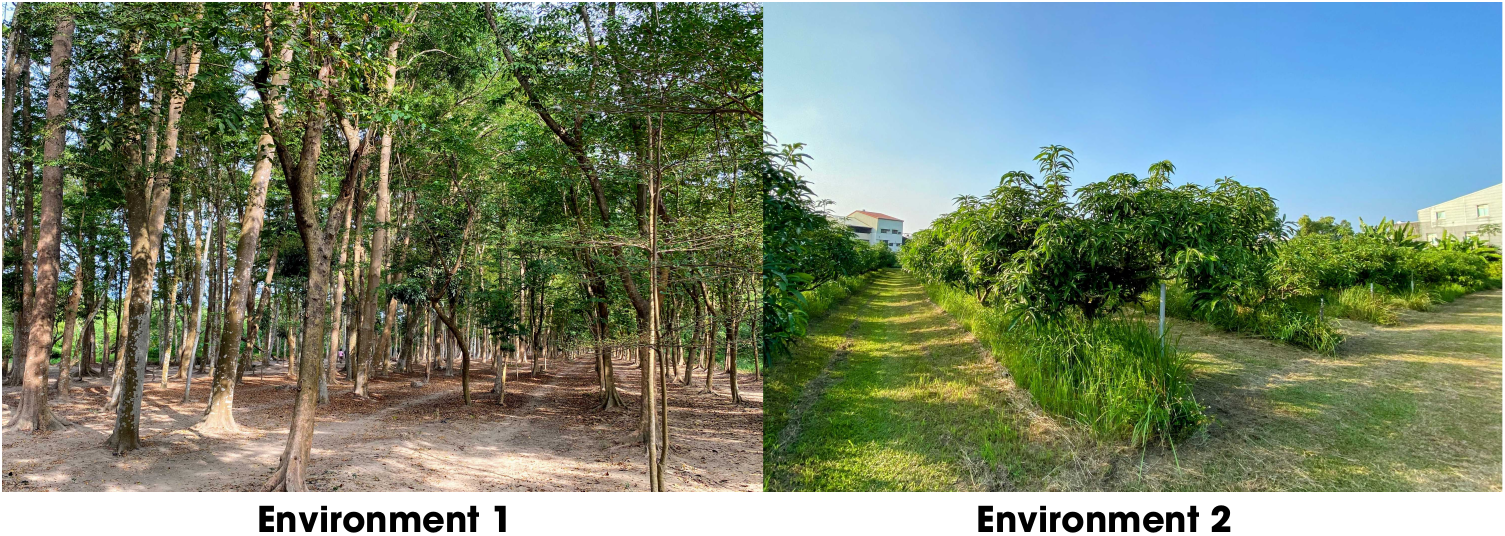}
\caption{Illustration of the two field environments used in the experiments.}
\label{fig:env_exp}
\end{figure}
\revcolor{The proposed system successfully constructs accurate global maps and generates precise paths that adapt to the distinct structural characteristics of different agricultural environments. As shown in Fig.~\ref{fig:global_path}, which presents the global map data and the corresponding paths generated for the two environments, the 3D LiDAR point cloud data effectively captured the characteristics of each environment.} In the first environment, the system accurately represents the non-uniform distribution of tree trunks, including cases where trunks are tilted or trees are misaligned within their respective rows. In the second environment, the mango orchard, the point cloud map reflects the real-world layout, featuring tree rows of varying lengths, an inconsistent number of trees per row, and dense canopy structures.

It is also notable that the detected tree positions align with their actual spatial distribution in the global map. In addition, the tree rows visualized using different colors were accurately identified using the VLM, demonstrating its practical potential for agricultural environments with complex and irregular tree distributions. Furthermore, by leveraging accurate row identification, the global paths generated along the crop rows align correctly with the rows that require spraying.

\revcolor{The VLM-based strategy effectively adapts the pesticide application rate to accommodate diverse wind conditions and crop densities, thereby minimizing chemical waste. Specifically, Fig.~\ref{fig:spraying_rate_env} illustrates the actual spray flow distribution of the proposed method  across two experimental environments under varying wind conditions}. Environment 1 comprises 497 trees arranged in 11 rows, whereas Environment 2 contains 204 trees distributed over 14 rows. The number of trees per row in Environment 1 varies considerably, ranging from 42 to 64 trees, while Environment 2 has a lower density, ranging from 11 to 17 trees per row. In the heatmap, the vertical axis (Row Index) represents the row number, and the horizontal axis (Plant Index Within Row) indicates the position of each tree within the row. The right column (Global Path Segments) depicts the corresponding global path segments for each row, \revcolor{ensuring that the robot covers both sides of the canopy to execute the adaptive spraying strategy effectively}. Color variations in the heatmap reflect adjustments in spray flow rates (L/min) according to wind direction and target tree positions, minimizing waste and maintaining uniform coverage.

The results show that the spray flow varies proportionally with wind intensity. At low wind speeds (2.0–2.3~m/s), the system maintains a stable balance between the two nozzles, with flow rates fluctuating slightly around 2.85–2.90~L/min due to minimal drift effects. In contrast, under stronger wind conditions (3.5–4.5~m/s), the amplitude of flow fluctuations increases significantly. Some regions exhibit higher flow, approximately 3.0~L/min (yellow), while other regions show lower flow, around 2.7~L/min (dark blue), forming alternating patterns across the field.

\begin{figure*}
\centering
\includegraphics[scale=0.398,page=1]{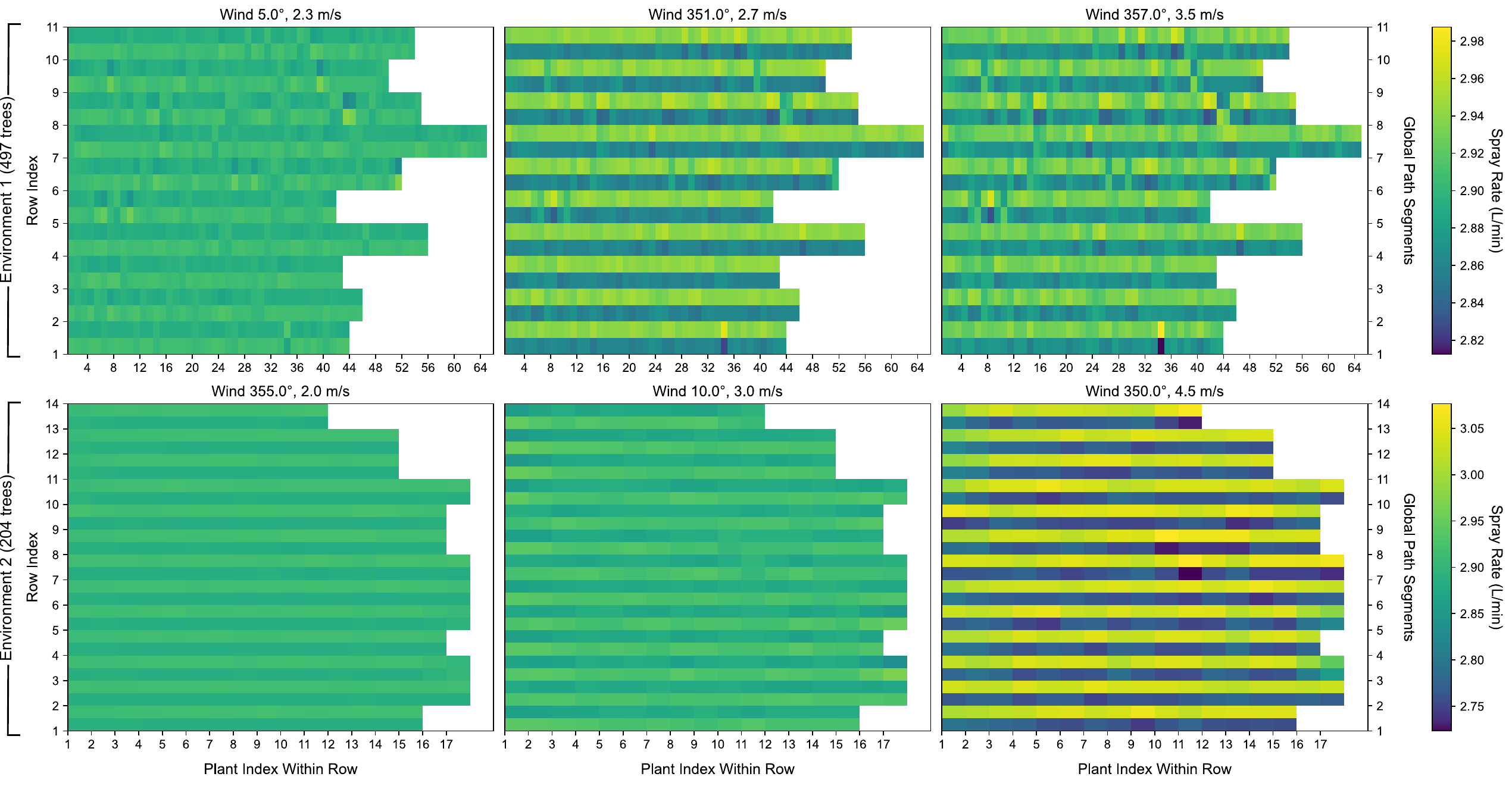}
\caption{Spray rate results in the two environments using the VLM-based spraying strategy. The robot adjusted the flow rate to reduce pesticide loss, particularly due to the influence of wind direction.}
\label{fig:spraying_rate_env}
\end{figure*}
\begin{figure}
\centering
\includegraphics[scale=0.435,page=1]{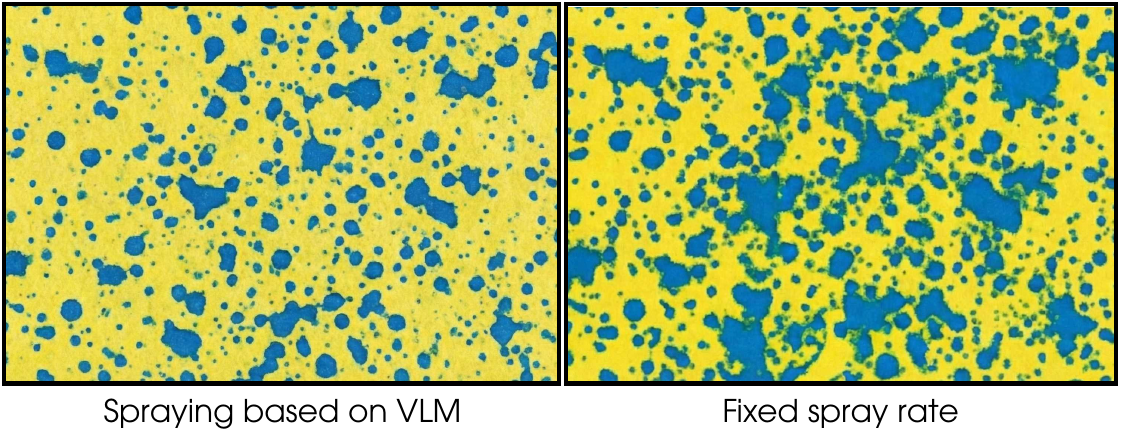}
\caption{Qualitative results of WSPs comparing the proposed VLM-based method with the baseline using a fixed pre-set spray flow rate when spraying in the windward direction.}
\label{fig:water_sensitive}
\end{figure}
Differences in the geometric structure of the two environments lead to distinct wind effects. Environment 1 exhibits relatively favorable aerodynamic conditions, whereas Environment 2, with a wind direction of $350^\circ$, experiences stronger crosswinds, resulting in pronounced pressure differences. The heatmap shows alternating color bands between adjacent tree rows, reflecting the robot's ability to adapt to a boustrophedon trajectory. A detailed examination indicates that even when moving against the wind, the robot does not apply a uniform flow rate across the entire row. Instead, the flow is dynamically adjusted according to the position and density of individual trees. These observations indicate that the VLM model effectively interprets the environmental context, enabling the robot to adjust its spraying strategy in a manner comparable to human reasoning.

\revcolor{The VLM-based spraying approach significantly improves spray uniformity and mitigates the risk of chemical runoff compared to conventional methods. This is visually supported by the enlarged Water Sensitive Papers (WSPs) samples presented in Fig.~\ref{fig:water_sensitive}, which were obtained when spraying in the downwind direction using both the proposed method and the fixed-rate baseline.} The samples sprayed with the fixed-rate method exhibit surface oversaturation with clustered liquid spots, indicating excessive chemical deposition and a high risk of runoff. In contrast, the VLM-based spraying approach generates a more uniform and discrete droplet distribution with adaptive coverage density, minimizing overlap. \revcolor{Moreover, the WSP samples were collected at various canopy locations and analyzed using ImageJ software \citep{schneider2012nih}. The analysis demonstrates that the VLM-based strategy achieves an average surface coverage of 26.4\% with an average droplet density of 62 droplets/cm\textsuperscript{2}, indicating highly uniform and adaptive deposition. Conversely, the fixed-rate baseline method results in an average surface coverage of 61.7\%, characterized by excessive droplet overlap and clustering, which significantly increases the risk of surface oversaturation and chemical runoff.}

\begin{figure*}
\centering
\includegraphics[scale=0.213,page=1]{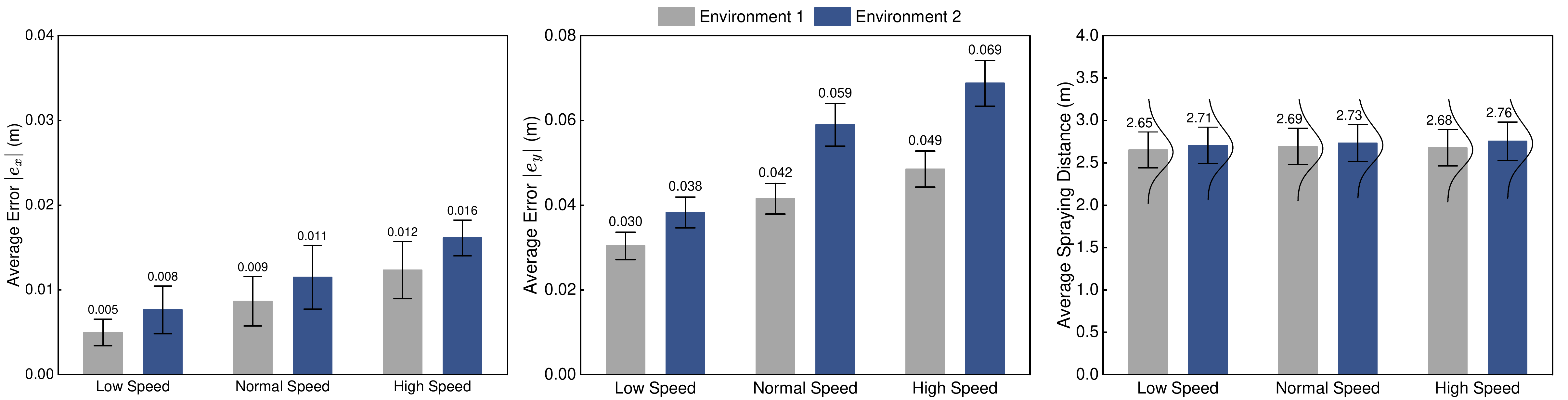}
\caption{Quantitative evaluation results of the controller at different velocities.}
\label{fig:tracking_error}
\end{figure*}
\begin{figure}
\centering
\includegraphics[scale=0.41,page=1]{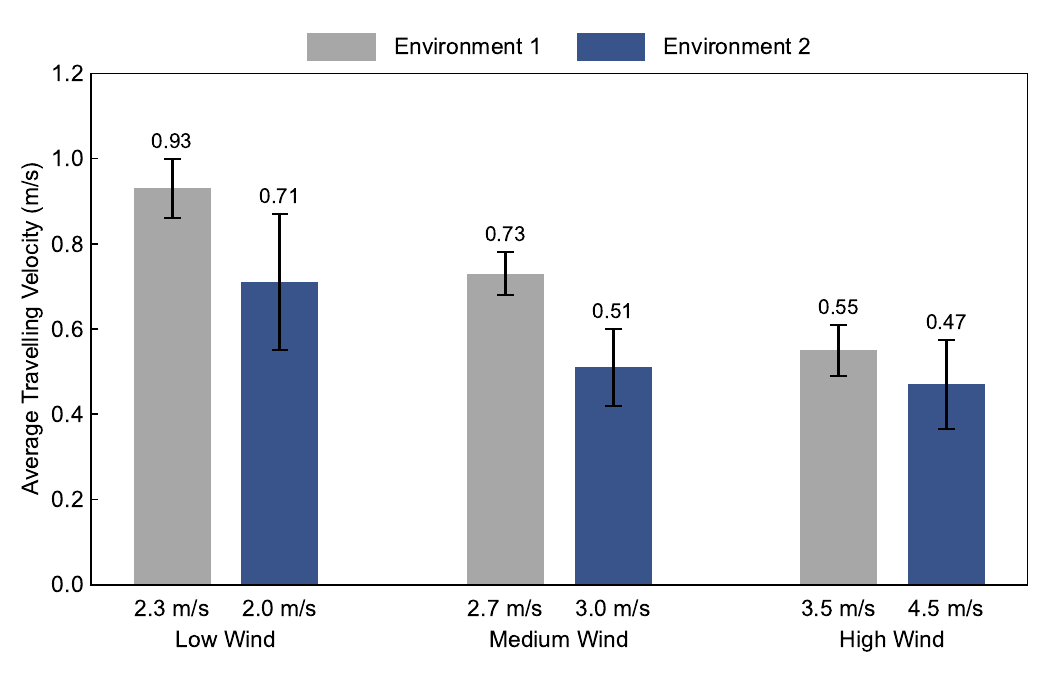}
\caption{Traveling speed of the robot in the two environments at three time points affected by different wind directions.}
\label{fig:spraying_vel}
\end{figure}
\revcolor{The proposed system proactively adjusts its traveling speed to maintain operational stability and ensure adaptive spraying effectiveness in response to changing environmental factors. As illustrated by the average traveling speed under varying wind conditions in Fig.~\ref{fig:spraying_vel}, the system employs an inversely proportional speed adjustment strategy with respect to wind intensity: as wind speed increases, the robot proactively reduces its speed to prioritize operational stability and spraying effectiveness.} Specifically, in Environment 1, the average speed decreases from 0.93~m/s under light wind (2.3~m/s) to 0.73~m/s under moderate wind (2.7~m/s), reaching a minimum of 0.55~m/s under strong wind (3.5~m/s). A similar trend is observed in Environment 2, although the overall speed range is lower, varying from 0.71~m/s to 0.47~m/s, reflecting differences in terrain structure between the two sites.

Furthermore, based on the VLM, the robot detects uneven terrain and dense vegetation in Environment 2, enabling human-like decision-making by maintaining a lower speed compared to Environment 1. The standard deviation also indicates that speed fluctuations in Environment 1 are smaller, reflecting smoother motion, whereas larger deviations in Environment 2 suggest that the robot must continuously adjust its speed to navigate terrain undulations.
\subsection{Controller Evaluation}
To ensure that the proposed controller operates effectively in agricultural environments with uneven terrain and dynamic conditions, particularly under the influence of sand, gravel, grass, and rough surfaces, we evaluated its performance at three different velocities: low speed (0.5 m/s), normal speed (1.0 m/s), and high speed (1.5 m/s). These tests were conducted to analyze tracking errors and to examine how travel speed affects crop detection and spraying decisions. The experiments were performed in the two environments described in Section \ref{sec:spraying_exp}. The MPPI controller demonstrates robust trajectory tracking performance and maintains low tracking errors across different operating speeds and terrains. As illustrated in Fig.~\ref{fig:tracking_error}, the detailed experimental results validate the robot's tracking accuracy and its consistent average spraying distance from the center to the tree.

The mean tracking errors along the $x$- and $y$-directions show that both the mean errors and their variabilities increase with operating speed. Notably, the lateral error $|e_y|$ is consistently higher than the longitudinal error $|e_x|$, reflecting the side-slip phenomenon commonly observed on agricultural terrains such as sand and grass. Moreover, the tracking errors in Environment~2 are consistently higher than those in Environment~1, indicating more nonlinear and less predictable wheel–terrain interactions. Specifically, in Environment~2, the lateral error $|e_y|$ increases from $3.8 \pm 0.36$ cm at low speed to $6.9 \pm 0.54$ cm at the highest speed (1.5 m/s). In contrast, Environment~1 exhibits a lower standard deviation, suggesting a more uniform surface condition. Additionally, the average spraying distance remains stable within the range of 2.65–2.76 m despite variations in speed and terrain. The differences in distance across operating speeds and environments are negligible, demonstrating that the system maintains accurate tree detection during motion and ensures consistent spraying performance under diverse conditions.

Although the tracking errors increase with the operating speed, the controller maintains a maximum error below 8~cm with a stable standard deviation, confirming the effectiveness and robustness of the proposed system in dynamic agricultural environments. At low and medium speeds, the tracking errors remain sufficiently small for precision spraying tasks, demonstrating that the controller not only ensures stability but also satisfies the accuracy requirements of practical agricultural operations.

A two-factor ANOVA was also conducted to evaluate the effects of operating speed and environment on system performance. Both factors significantly affected the lateral error $|e_y|$ ($p < 0.05$), while the longitudinal error $|e_x|$ was mainly influenced by speed, with a smaller contribution from the environment. For spraying distance, differences were statistically significant ($p < 0.05$), but the effect of speed was minor and mainly dependent on the terrain characteristics. 
\subsection{Discussion}
\revcolor{
\subsubsection{Performance of VLMs in Crop-Row Detection}
The experimental results highlight the effectiveness and practical potential of VLMs in agricultural robotics. By leveraging their capability to interpret visual information and reason about contextual relationships, the proposed approach achieves substantially higher detection accuracy than conventional methods. The system can identify crop rows and infer their spatial structure in a manner comparable to human visual perception. This capability is further demonstrated through its integration into the robotic spraying strategy.

A key finding of this study is that the VLM-based approach improves detection accuracy by approximately 30\% to 40\% compared with traditional unsupervised clustering methods such as K-means and DBSCAN \citep{10773736, s23104808}. Conventional geometric approaches rely primarily on spatial point distributions and are therefore susceptible to environmental noise, including weeds, occlusions, and overlapping canopies. In contrast, the VLM is able to infer the underlying crop-row structure through semantic understanding of the scene. As a result, the proposed method maintains robust performance in complex agricultural environments while avoiding the extensive parameter tuning commonly required by heuristic-based approaches \citep{10604355}.
\subsubsection{VLM-Driven Context-Aware Spraying Strategy}
The incorporation of VLMs into the spraying strategy enables adaptive and context-aware pesticide application. Unlike conventional detect-and-spray systems that depend on instantaneous sensor-triggered actions, the proposed framework considers multiple environmental factors, including wind direction, foliage density, and terrain conditions, when generating spraying decisions. This capability allows the robot to adjust pesticide flow rates and operating speed according to the surrounding context. Consequently, chemical waste can be reduced while maintaining consistent spraying coverage under varying field conditions.

Furthermore, we evaluated the robustness of the VLM-based decision-making process against variations in input prompts. Our experimental testing demonstrates that the system maintains high stability in spraying decisions; slight variations in prompt phrasing, such as reordering descriptive adjectives or modifying sentence structures while preserving the core intent, do not significantly alter the output speed and flow rate, provided that the environmental parameters remain consistent. This stability is primarily attributed to Few-Shot and CoT prompting strategies, which guide the model to prioritize semantic interpretation over superficial syntax, as well as the contextual understanding capability of large language models. However, to ensure system-level reliability, restricting the output format to be compatible with the system is necessary. This ensures that the robot executes only valid, hardware-compatible commands, thereby mitigating potential errors arising from model outputs.

Although large-scale VLMs generally exhibit higher computational latency than conventional algorithms, the experimental results indicate that this limitation does not hinder practical field deployment. Agricultural robots typically operate at relatively low speeds, allowing sufficient time for high-level perception and decision-making processes. In the proposed framework, the VLM performs global scene understanding and strategic planning at a lower update frequency, whereas the MPPI controller handles trajectory tracking and collision avoidance at a higher control frequency. This hierarchical architecture enables the system to satisfy real-time operational requirements while benefiting from the advanced reasoning capabilities of the VLM. Nevertheless, further reductions in inference latency through model compression, quantization, and edge deployment remain important directions for future research, particularly for applications that require faster reactive responses.
\subsubsection{System Limitations and Future Work}
Despite the promising results, this study has certain limitations that must be acknowledged. First, the proposed framework was evaluated in a restricted number of field environments, test farms, and geographical locations. Because agricultural conditions, canopy structures, and microclimates vary significantly across different regions, further extensive testing in more diverse and unstructured agricultural settings is necessary to fully validate the system's generalizability.

Second, field experiments indicate that controller performance remains sensitive to terrain variability. In the mango orchard environment, nonlinear wheel–terrain interactions contributed to increased lateral tracking errors. These results suggest that while the VLM provides effective high-level planning and decision-making capabilities, overall system performance is still influenced by low-level locomotion challenges associated with uneven agricultural terrain. Future research should therefore investigate terrain-aware trajectory optimization methods that favor smoother traversable regions. In addition, hardware improvements, such as chassis designs incorporating passive stabilization or enhanced suspension mechanisms, may further improve operational robustness in rugged field environments.

Finally, due to institutional constraints and ongoing proprietary research, the complete source code and full datasets cannot be publicly released at this time. However, to support replicability and allow other researchers to build upon this work, detailed pseudocode representing the core methodology has been provided and can be found in the Data Availability section.}
\section{Conclusion}\label{sec:conclusion}
\revcolor{
In this study, we presented a novel context-aware pesticide spraying framework for agricultural robots powered by VLMs. By leveraging the human-like reasoning capabilities of VLMs, the proposed system directly recognizes crop rows from visual inputs and generates adaptive spraying strategies. These strategies are formulated by dynamically synthesizing real-time imagery with critical contextual factors, including weather conditions, time, pesticide type, geographical location, and crop variety. Consequently, the robot selects adaptive spraying parameters tailored to specific environmental conditions and crop growth stages, thereby minimizing pesticide waste and reducing adverse environmental impacts. Additionally, an MPPI controller was implemented to ensure precise and safe trajectory tracking during field operations. Experimental evaluations conducted in two distinct agricultural environments demonstrated that our approach improves crop-row detection accuracy by at least 30\% compared to conventional methods. The results also confirmed the system's ability to make adaptive, human-like spraying decisions under varying conditions. Future research will focus on validating the framework in more complex, unstructured agricultural environments and deploying VLMs locally on edge devices to reduce inference latency and achieve high-frequency real-time performance.}
\section*{Declaration of Generative AI and AI-assisted technologies in the writing process}
During the preparation of this work, the authors used ChatGPT and Gemini to improve the readability of the manuscript. After using these tools, the authors reviewed and edited the content as needed and take full responsibility for the final manuscript.
\section*{CRediT authorship contribution statement}
\textbf{Cong-Thanh Vu:} Writing - original draft, Methodology, Investigation, Formal analysis, Data curation, Conceptualization. \textbf{Yen-Chen Liu:} Writing-review \& editing, Supervision, Funding acquisition, Conceptualization.
{\section*{Declaration of competing interest}
The authors declare that they have no known competing financial interests or personal relationships that could have influenced the work reported in this paper.
\section*{Acknowledgments}
This work was supported in part by the National Science and Technology Council (NSTC), Taiwan, under Grants NSTC 114-2622-E-006-011 and NSTC 114-2218-E-006-021.
\bibliographystyle{cas-model2-names}

\bibliography{cas-refs}


\end{document}